\documentclass[10pt,twocolumn,letterpaper]{article}

\usepackage{iccv}
\usepackage{times}
\usepackage{epsfig}
\usepackage{graphicx}
\usepackage{amsmath}
\usepackage{amssymb}

\usepackage{subcaption}
\usepackage{amsmath}
\usepackage{multirow}

\DeclareMathOperator*{\argmin}{arg\,min}

\usepackage[pagebackref=true,breaklinks=true,letterpaper=true,colorlinks,bookmarks=false]{hyperref}

\usepackage[breaklinks=true,bookmarks=false]{hyperref}

\iccvfinalcopy 

\def\iccvPaperID{****} 

\ificcvfinal\fi

\newcommand{\HW}[1]{{\color{black}#1}}
\newcommand{\HWW}[1]{{\color{black}#1}}

\begin{document}

\title{Graph Constrained Data Representation Learning for \\ Human Motion Segmentation \vspace{-1mm}}


\author{%
Mariella Dimiccoli$^1$ \quad Llu\' is Garrido$^2$ \quad Guillem Rodriguez-Corominas$^1$ \quad Herwig Wendt$^3$\\
\normalsize
$^1$Institut de Rob\'otica i Inform\'atica Industrial (CSIC-UPC) \quad $^2$Univ. of Barcelona \quad $^3$CNRS, IRIT, Univ. of Toulouse  \\
\qquad{\footnotesize {\tt \{mdimiccoli,grodriguez\}@iri.upc.edu} \qquad\qquad\quad  {\tt lluis.garrido@ub.edu} \quad\qquad {\tt herwig.wendt@irit.fr}}\vspace{-3mm}
}

\maketitle
\ificcvfinal\thispagestyle{empty}\fi


\begin{abstract}
Recently, transfer subspace learning based approaches have shown to be a valid alternative to unsupervised subspace clustering and temporal data clustering  for human motion segmentation (HMS).
These approaches leverage prior knowledge from a source domain to improve clustering performance on a target domain, and currently they represent the state of the art in HMS.
Bucking this trend, in this paper, we propose a novel unsupervised model that learns a representation of the data and digs clustering information from the data itself. Our model is reminiscent of temporal subspace clustering, but presents two critical differences. 
First, we learn an auxiliary data matrix that can deviate from the initial data, hence confers more degrees of freedom to the coding matrix. 
Second, we introduce a regularization term for this auxiliary data matrix that preserves the local geometrical structure present in the high-dimensional space.
The proposed model is efficiently optimized by using an original Alternating Direction Method of Multipliers (ADMM) formulation allowing to learn jointly the auxiliary data representation, a nonnegative dictionary and a coding matrix. 
Experimental results on four benchmark datasets for HMS demonstrate that our approach achieves significantly better clustering performance then state-of-the-art methods, including both unsupervised and more recent semi-supervised transfer learning approaches\footnote{Project page at URL \hyperlink{https://github.com/mdimiccoli/GCRL-for-HMS/}{https://github.com/mdimiccoli/GCRL-for-HMS/}.}. 
\end{abstract}

\vspace{-5mm}

\section{Introduction}

Human-motion segmentation (HMS) aims at breaking a continuous sequence of data depicting human actions and activities into a set of internally coherent temporal segments, and has emerged as a suitable first step in early processing of untrimmed videos for human action recognition \cite{jhuang2013towards}. 
Despite the fact that HMS has been intensively investigated so far \cite{tierney2014subspace,li2015temporal,xia2017human}, performance have not yet reached the level of 
accuracy required in real world applications.

Because labelling large amounts of videos for creating a training set for HMS is very expensive and time-consuming, the problem has been traditionally addressed through unsupervised learning techniques  \cite{oh2008learning,barbivc2004segmenting,lv2006recognition}. 
Several approaches cast HMS as a clustering problem by relying on the framework of subspace clustering \cite{vidal2011subspace}. The key idea of subspace clustering is to learn an effective representation in form of a coding matrix from which it is possible to construct an affinity matrix that allows to separate data points according to their underlying low-dimensional subspaces. These subspaces are assumed to correspond with the different motions.
However, these clustering approaches typically do not take into account temporal continuity and, moreover, they are sensitive to noise.
Some recent subspace clustering based approaches have focused on dealing with noise and outliers naturally present in the data, as well as  with its sequential nature \cite{elhamifar2009sparse,elhamifar2013sparse,tierney2014subspace,li2015temporal,li2019robust}.  Outside the subspace clustering framework, other temporal clustering approaches work directly on the temporal data instead of on an underlying representation to estimate cluster labels \cite{smyth1999probabilistic,xiong2002mixtures,keogh2001locally}.
Nevertheless, the main challenge of all these unsupervised approaches remains how to cope with the lack of prior knowledge, which could easily cause unpredictable segmentation output. Recently, this problem has been partially addressed by transfer subspace techniques, that leverage prior information from source data to improve clustering accuracy on the target data \cite{wang2018learning,wang2018low,zhou2020multi}. These methods have indeed reported improved performance and currently represent the state of the art in HMS. 
\\
\indent Instead of leveraging information from source annotated data, here we propose to cope with the lack of prior knowledge in an unsupervised fashion by further exploiting the local geometric structure present in the original high-dimensional space.
Our \textit{a priori} on the data samples is that they are drawn from an unknown underlying similarity graph, where nodes in the graph correspond to individual frames, edges to connections between nodes, and communities (i.e., groups of nodes that are interconnected by edges with large weights) roughly correspond to different motions. Our model aims at jointly learning the subspaces and a compatible representation for the similarity graph. Since we are interested in pushing the limits of state-of-the-art HMS, we build on the framework of Temporal Subspace Clustering (TSC) \cite{li2015temporal}, that has proven to be the most effective subspace clustering technique for temporal data without transfer. The rational underlying TSC is to find a coding matrix whose affinity graph leads to good clustering with temporal coherence. This is ensured by a least squares regression formulation where the original data are approximated through a nonnegative dictionary and code under block diagonal and temporal Laplacian regularization terms that lend the model its global subspace structure and temporal smoothness, respectively. 
However, when the original data is noisy or corrupted, nuisances propagate also to the coding matrix. To avoid this, we allow the auxiliary data representation to vary, hence conferring to the coding matrix more degrees of freedom and robustness to subspace assumption violations and noise. Our prior for the auxiliary data matrix is that its columns act as representation vectors for the affinity matrix (graph)  of  the  original  data. Therefore, we introduce a graph regularization term, that controls the distance between the affinity graphs  of the original and auxiliary data. This effectively allows the auxiliary data to cope with data nuisances while preserving the local geometrical structure of the original data.
We propose an efficient optimization formulation for our model, based on the  Alternating Direction Method of Multipliers (ADMM).

Our main contributions are summarized as follows: 1) we propose a novel approach for HMS, which jointly learns an auxiliary data matrix, a non-negative dictionary and a coding matrix under temporal, block-diagonal and graph constraints via an original ADMM formulation, 2) we introduce an original graph regularization term that preserves the geometrical structure of the original data in high-dimensional space, 3) we present a comprehensive analysis of our model, including ablation study and sensitiveness analysis, 
4) we achieve significant performance improvements (up to $\approx 20\%$ accuracy and $\approx 5\%$ NMI) over the state-of-the-art methods on four public benchmarks for HMS.

 \section{Related work}

\noindent{\bf Subspace clustering.} Subspace clustering assumes sampled data to be drawn from a union of multiple subspaces, hence breaking the assumption that all of the clusters in a dataset are found in the same (sub)set of dimensions. Under this assumption, the sampled data obey the so called self-expressiveness property, i.e., each data point in a union of subspaces can be represented by a linear combination of other points in the dataset. This can be formulated as
$$X = XZ,$$
where $Z \in \mathbb{R}^{r \times N}
$ is the representation coefficient or coding matrix.
The goal of subspace clustering is to find a coding matrix $Z$ that minimizes the reconstruction error under some regularity constraints. The final clusters are obtained by applying spectral clustering to the affinity matrix of $Z$. Different variants of subspace clustering have been proposed that impose different constraints on $Z$ for this purpose.
Least Square Regression (LSR) \cite{lu2012robust} uses an $\ell_2$ norm regularizer for $Z$. Low-rank representations (LRR) \cite{liu2012robust} impose on $Z$ to have small rank and are therefore more robust to corrupted data. Sparse subspace clustering (SSC) algorithms \cite{elhamifar2009sparse} enforce a sparsity constraint on $Z$ and are also applicable to noisy data that do not cluster perfectly into subspaces.
Probabilistic methods usually model the data points using a mixture of probabilistic PCAs and are often more
robust to noise and outliers than LSR and LRR \cite{babacan2012probabilistic}. Information theoretic learning based framework such as \cite{li2019robust} have proved to be robust to piecewise identically distributed noise. 
One of the shortcomings of all of the above methods is that they neglect the temporal information that is  implicitly encoded in sequential data. 

\noindent{\bf Temporal data clustering.}
Temporal data clustering methods explore dynamic regularities underlying temporal data in an unsupervised learning fashion.
Following the taxonomy introduced in \cite{yang2010temporal}, we can distinguish three general approaches to 
temporal data clustering, depending on the data dependency treatment: model-based \cite{smyth1999probabilistic,xiong2002mixtures}, temporal proximity based \cite{keogh2003need} and representation based algorithms \cite{dimitrova1995motion,chen1999motion,keogh2001locally}.
Model-based and temporal proximity based approaches work directly on temporal data and deal with the temporal correlation during clustering analysis. A representative example in the context of HMS is Aligned Clustering analysis (ACA) \cite{zhou2012hierarchical}, that extends kernel k-means and spectral clustering for time series  clustering, formulating the problem within an optimization framework.

Instead, representation based approaches capture temporal data dependencies via a sparse representation. Within this class fall all methods that tried to adapt the subspace clustering framework \cite{elhamifar2013sparse} to sequential data.
Tierney et al. \cite{tierney2014subspace} proposed an Ordered Subspace Clustering (OSC)
method that incorporates a neighbour penalty term to enforce consecutive frames to have similar representation.
In Sequential Subspace Clustering   \cite{wu2015ordered} a quadratic normalizer is imposed on
the sparse coefficients to model the temporal correlation
among the data points. Clustering accuracy is enhanced by incorporating into the model  a block-diagonal prior for the spectral clustering affinity matrix.
Temporal Subspace clustering (TSC) \cite{li2015temporal} learns jointly a dictionary and a representation code with a Laplacian temporal regularization. The use of a dictionary allows for more expressive coding.
These temporal clustering methods cannot explicitly and efficiently deal with corrupted or partially missing data.
A probabilistic approach to TSC was proposed in  \cite{gholami2017probabilistic}, that models temporal dependencies by using Gaussian Process (GP) priors \cite{seeger2004gaussian} and is more robust to outliers.

\noindent{\bf {Transfer subspace-learning temporal clustering.}}
Unsupervised subspace learning methods work without any prior knowledge and achieve reasonable but relatively modest performance.
Recently, transfer subspace learning-based approaches \cite{wang2018learning,wang2018low,zhou2020multi} have reported improved performance in motion segmentation. These methods adapt to the temporal data clustering problem the concept of transfer learning \cite{pan2009survey}. Transfer learning, also commonly know as domain adaptation, aims at transferring knowledge from a source domain to a target domain, typically by  imposing the data distributions of two domains to be similar in a semi-supervised fashion. Contrary to \cite{wang2018learning,wang2018low}, that carry out the transfer subspace learning on the original high-dimensional feature space,  \cite{zhou2020multi}  factorizes the original features of the source and the target data into implicit multi-layer feature spaces and uses them to fuse multi-level structural information effectively.

 \begin{figure}[t]
\begin{center}
   \includegraphics[width=\linewidth]{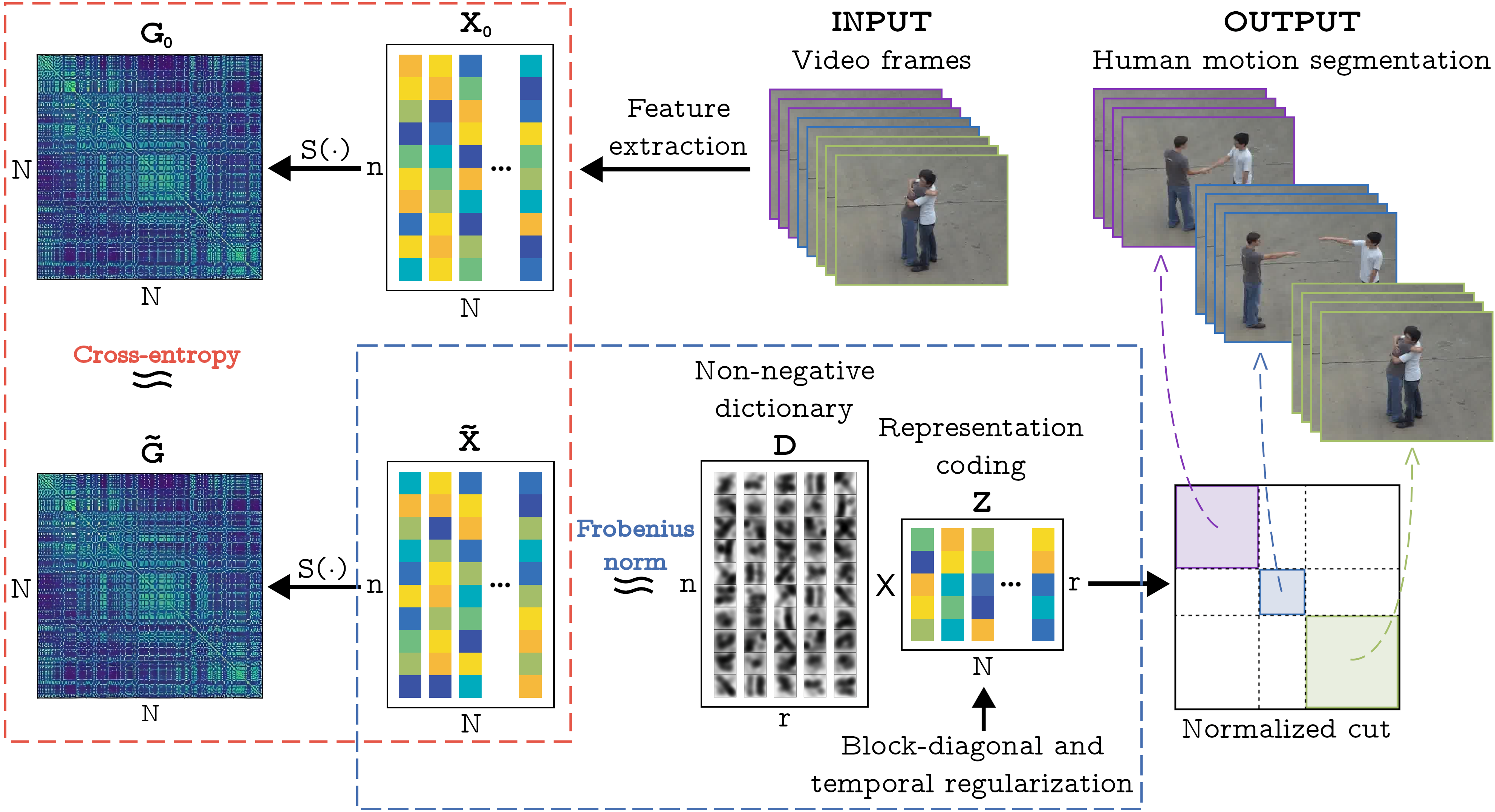}
\end{center} 
\vspace{-5mm}
   \caption{Overview of the proposed framework for HMS.}
\label{fig:overview}
\vspace{-2mm}
\end{figure}

\section{Approach}
\label{sec:approach}
Fig.~\ref{fig:overview} provides an overview of our approach for HMS. Our method take as input frame-level features extracted from the video sequence, and firstly computes an affinity graph $G_0$ in the original high-dimensional feature space.  Then, it optimizes jointly a non-negative dictionary $D$ and coding matrix $Z$, and the auxiliary data representation $\tilde{X}$ under temporal, block-diagonal and graph based regularization constraints. The final segmentation is obtained by applying normalized cut on the affinity graph of $Z$. 

\subsection{Problem formulation}
\noindent{\bf Preliminaries.}
Let us denote $X = [x_1,x_2,...,x_N] \in \mathbb{R}_+^{n \times N}$ time-series data corresponding to motion features with $N$ time-steps in an $n$-dimensional Euclidean space. 
Without loss of generality, we assume that $X$ is positive valued and normalized to the interval $[0,1]$. Furthermore, we assume that $X$ is drawn from a union of $M$ subspaces $\{\phi_m\}_{m=1}^M$ of unknown dimensions $dim(\phi_m) =M_m(0 < M_m < n)$, and aim at grouping the $N$ vectors into $p(p \geq M)$ sequential motion segments, and cluster these $p$ segments into their respective subspaces.

Given a nonnegative dictionary (i.e., set of atoms) $D = [d_1,d_2,\ldots,d_r] \in \mathbb{R}_+^{n\times r}$ and a coding matrix $Z \in \mathbb{R}_+^{r\times N}$, the time series data can be
approximately represented as 
$$X \approx DZ,$$ 
where  $r$ is the number of atoms in the dictionary. 
The use of the dictionary in the above formulation instead of the self-representation model (i.e., $X \approx XZ$) allows to learn a more expressive coding matrix $Z$ from the data \cite{li2015temporal}. $Z$ can then be used to construct an affinity graph for subspace clustering, and the temporal clustering results can be obtained by applying an efficient clustering algorithm, e.g. normalized cuts.
Yet, noisy or corrupted data samples contained in $X$ as well as portions of temporal segments that deviate from the subspace assumption may nonetheless propagate to $Z$ and $D$ and lead the model astray.

\noindent{\bf Graph constrained data representation.}
Our key insight is therefore to let $DZ$ approximate auxiliary data $\tilde X \in \mathbb{R}_+^{n \times N}$ instead of $X$, that we optimize jointly with $D$ and $Z$. This lends the model extra flexibility to find more expressive codes $Z$. 
Our prior for the auxiliary data $\tilde X$ is that they act as representation vectors for the affinity matrix (graph) of the original data $X$. We additionally assume that communities in the similarity graph roughly correspond to subspaces, where nodes correspond to individual frames and edges correspond to similarity values between frames. Thus, the auxiliary data points $\tilde X$ approximately reproduce the local geometry of the original data points $X$ but can otherwise move in $\mathbb{R}_+^{n\times N}$ to find a representation that is more compatible with the subspace clustering model, and to unburden $Z$ and  $D$ from dealing with corrupted data points.

The assumptions on $\tilde X$ are expressed through a regularization term that is reminiscent of graph embedding \cite{goyal2018graph} and penalizes deviations of the graph of the new data points $\tilde X$ from the graph of $X$.
Specifically, the graphs of $X$ and $\tilde X$ are represented by their respective $ \mathbb{R}^{N \times N}$ affinity matrices ${G}_0 =S(X)$ and ${\tilde G} =S(\tilde X)$. 
These are defined element-wise via a pairwise similarity, upon normalization to unit total weight,
\begin{multline}
\label{equ:cosdist}
(S(X))_{kj}=w_{kj}/W,\quad W = {\sum}_{(k,j) \in E} w_{kj}\\
w_{kj} = \exp\big(-(1-d(X(j),X(k)))/h\big),
\end{multline} 
where $E$ is the set of pairwise relations, $d(\cdot,\cdot)$ is the cosine distance and $h$ the filtering parameter of the exponential function.
%
%
The normalized pairwise similarities can be interpreted as empirical joint probability distributions between pairs of vertices $\hat{p}(k,j)$. 
Therefore,  Eq.~\eqref{equ:cosdist} defines a distribution $p(\cdot,\cdot)$ over the space $N\times N$.
Then, to preserve first-order neighbor relations, we minimize the KL-divergence of the two joint probability distributions $\hat{p}_{G_0}(\cdot,\cdot)$, $p_{\tilde{G}}(\cdot,\cdot )$ that correspond with the original and learnt auxiliary data graphs, respectively. Upon removal of constant terms, this yields the cross entropy (CE) divergence CE$(\hat{p}_{G_0}(\cdot,\cdot), p_{\tilde{G}}(\cdot,\cdot ))= -\sum_{(k,j)} \hat{p}_{G_0}(k,j)\log(p_{\tilde{G}}(k,j))$; we denote
this additional loss term as $\mathcal{L}_G$ and name it graph constrained loss.
Our CE term is easy to minimize and strongly convex, guaranteeing convergence of our algorithm, see Sec. \ref{sec:modelstudy}.

\noindent{\bf Model.}
With this graph constrained loss and the new data fidelity term on $\tilde{X}$, we obtain our final model:
\begin{multline}
\label{equ:SC0}
\min_{Z,D,\tilde X} \mathcal{L}_G (S(\tilde{X}),{G}_0) + \lambda_0||\tilde X-DZ||_F^2 
\\
+ \lambda_1||Z||_F^2+\lambda_2f(Z)\\
\qquad s.t. \quad Z\geq 0,\;D\geq 0,\;||d_i||_2^2\leq 1, \;i=1,\ldots,r.
\end{multline}
The Frobenius norm on $Z$ is used to enforce block diagonal structure in $Z$, see e.g. \cite{lu2018subspace}. Moreover, $f(Z)=\mbox{tr}(Z^TL_TZ)$ is a weighted Laplacian $\ell^2$ regularization for temporal consistency defined as in \cite{li2015temporal}. 
The unit norm constraints on $d_i$ resolve the scale ambiguity between $D$ and $Z$.

\subsection{Optimization}
To solve the objective function in \eqref{equ:SC0}, we devise an optimization algorithm based on ADMM. 
Upon introduction of a set of auxiliary variables $Y$, $U$ and $V$, \eqref{equ:SC0} is formulated equivalently as
\begin{align*}
\min_{\tilde X, U,V,Y,D,Z}  
& \mathcal{L}_G (S(\tilde{X}),{G}_0) \\
& + \lambda_0||Y-UV||_F^2 + \lambda_1||V||_F^2+\lambda_2f(V)\\
s.t. & \quad Y = \tilde X,\quad U = D,\quad V = Z,\\
& \quad Z\geq 0,\;D\geq 0,\;||d_i||_2^2\leq 1, \;i=1,\ldots,r.
\end{align*}
The augmented Lagrangian for this problem is
\begin{align}
L_{\rho} =  
&  \mathcal{L}_G (S(\tilde{X}),{G}_0) 
\nonumber
\\
& + \lambda_0||Y-UV||_F^2+ \lambda_1||V||_F^2+\lambda_2\textnormal{tr}(VL_TV)
\nonumber
\\
& + \langle\Lambda_{\tilde X}, Y-\tilde X\rangle
+ \langle\Lambda_U, U- D\rangle
+ \langle\Lambda_V, V- Z\rangle
\nonumber
\\
& + \frac{\rho}{2}|| Y - \tilde X||_F^2
+ \frac{\rho}{2}|| U-D||_F^2
+ \frac{\rho}{2}|| V-Z||_F^2
\nonumber
\\
s.t.& \quad Z\geq 0,\;D\geq 0,\;||d_i||_2^2\leq 1, \;i=1,\ldots,r.
\label{equ:augLagNew0}
\end{align}
The ADMM algorithm that solves this optimization problem \eqref{equ:augLagNew0} is given by alternately minimizing $L_\rho$ w.r.t. each variable $Y,V,U,\tilde X,Z,D$ individually \cite{boyd2011distributed}.

\noindent{\bf Update for $V,\,U$.\quad}  
The updates for $V,\,U$ are similar to those of the TSC algorithm of Li et al. \cite{li2015temporal}, but need to be slightly modified to account for the new term $\mathcal{L}_G$ and the additional regularization parameter $\lambda_0$. Specifically, the update for $V$ is obtained as the solution to the Sylvester equation
\begin{multline}
\label{equ:update:V}
{(2\lambda_0 U^TU + \lambda_1 I +\rho I)} V + \lambda_2 V L_T \\= {2\lambda_0 U^TY -\Lambda_U + \rho Z}
\end{multline}
and the update for $U$ has the closed-form solution
\begin{equation*}
U = (2\lambda_0 YV^T - \Lambda_U + \rho D )(2\lambda_0  VV^T + \rho I)^{-1}.
\end{equation*}

\noindent{\bf Update for $Z,\,D$.\quad}  
The updates for $V,\,U$ are identical to those of the TSC algorithm of Li et al. \cite{li2015temporal} and are given by the expressions
\begin{align*}
Z &= \mathcal{F}_+(V + \Lambda_V/\rho),\\
D &= \mathcal{F}_+(U + \Lambda_U/\rho),
\end{align*}
where $(\mathcal{F}_+(A))_{ij}=\max(A_{ij},0)$.

\noindent{\bf Update for $Y$.\quad} 
We cancel the gradient of \eqref{equ:augLagNew0} w.r.t. $Y$
\begin{multline*}
\nabla_{ Y} L_{\rho} = 
2\lambda_0Y-2\lambda_0UV  + \Lambda_{\tilde X} +\rho(Y -\tilde X)
\end{multline*}
and find the update for $Y$
$$
Y = (2\lambda_0+\rho)^{-1}(2\lambda_0UV - \Lambda_{\tilde X} +\rho\tilde X).
$$

\noindent{\bf Update for $\tilde X$.\quad}  
The update for $\tilde X$ does not have a closed form solution since one has to solve
$$\tilde{X} \!=\! \argmin_{\tilde{X}} \mathcal{L}_G (S(\tilde{X}),G_0) + \langle\Lambda_{\tilde X}, Y-\tilde X\rangle + \frac{\rho}{2}|| Y - \tilde X||_F^2.
$$
This expression is easy to solve using gradient descent.
The expression for the gradient 
is given by
$$
\nabla_{\tilde X} L_{\rho} =  \nabla_{\tilde X}\;  \mathcal{L}_G (S(\tilde{X}),{G}_0)  - \Lambda_{\tilde X} - \rho(Y- \tilde X),
$$
and thus essentially requires the calculation of the gradient of the cross-entropy loss $ \mathcal{L}_G$ composed with the similarity function $S$.
\HW{Both $Y$ and $\tilde X$ are normalized to $[0,1]$, as $X$ is assumed to be.}

\noindent{\bf Update for the Lagrange multipliers $\Lambda$.\quad} Finally, the Lagrange multipliers are updated as usual as
\begin{align*}
\Lambda_{U}& \gets \Lambda_{U} + \rho(U-D), \\
\Lambda_{V}& \gets \Lambda_{V} + \rho(V-Z), \\
\Lambda_{\tilde X}& \gets \Lambda_{\tilde X} + \rho( Y -\tilde X).
\end{align*}

The details for solving Eq. \eqref{equ:augLagNew0} via the above ADMM formulation are summarized in Algorithm 1. 
Note that the normalization of $X$ and $\tilde X$,
in addition to the strong convexity property of the \HWW{CE function}, allows to ensure mild convergence conditions for the graph regularization term, see Sec.~\ref{sec:modelstudy}.

\begin{figure}[h]
\includegraphics[width=0.99\linewidth]{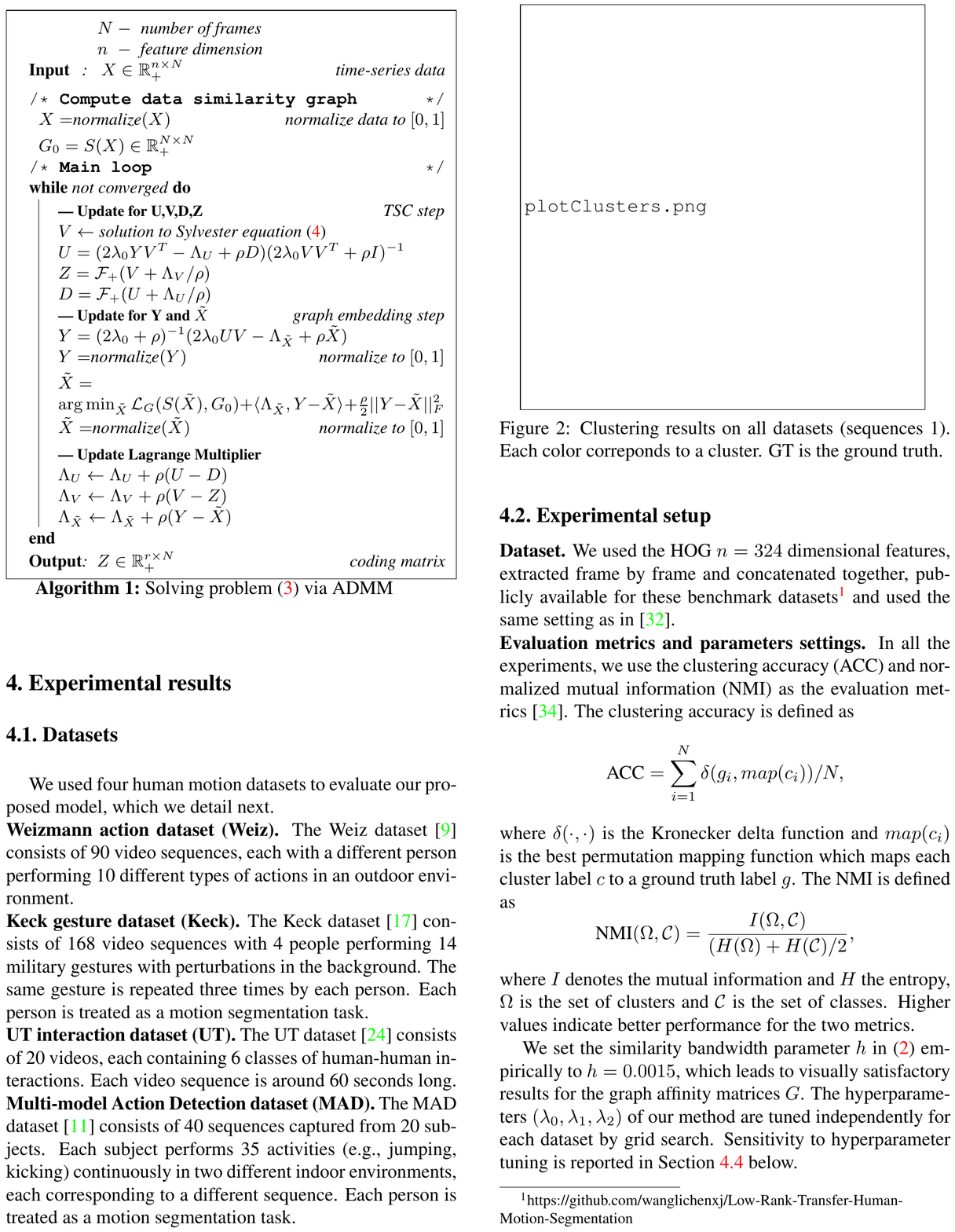}
\vspace{-2mm}
\end{figure}

\subsection{Clustering}
To produce the temporal segmentation results, we applied the Normalized Cut algorithm  \cite{shi1998motion} to the affinity matrix $A$ of the learnt coding matrix $Z$. We proceed as in the TSC algorithm \cite{li2015temporal} and define each element of $A$ as the cosine distance between the respective pair of the learnt code vectors
\begin{equation}
\label{equ:simZ}
    (A(Z))_{kj} = \frac{z_j^T z_k}{||z_j^T||_2||z_k||_2}.
\end{equation}

\begin{table*}
\centering
\caption{Clustering comparison results in terms of  ACC and NMI on four HMS datasets. For the transfer subpaces learning approaches (TSS, LTS and MTS), we reported the best results and the source dataset used to obtain them between parenthesis (M,K, U, and W stands for MAD, Keck, UT, and Weiz respectively). The best clustering results are denoted in bold.}
    \begin{subtable}{0.50\columnwidth}
    \centering
    \noindent\setlength{\tabcolsep}{1pt}
\begin{tabular}{c|r|rr} 
\hline 
\centering
Method & ACC & NMI \\ [0.5ex]
\hline 
\centering
LRR & 0.4382 & 0.3638\\
RSC & 0.4112  &  0.4894\\ 
OSC & 0.7047 & 0.5216\\
SSC & 0.6009 & 0.4576\\
LSR & 0.5093 & 0.5091\\
TSC & 0.6111 & 0.8199 \\ 
\hline
TSS(M) & 0.6208 & 0.8509 \\
LTS(K) & 0.6391 & 0.8599 \\ 
MTS(K) & 0.6436 & 0.8371 \\
\hline
\bf Ours & ${\bf 0.8501}$ & ${\bf 0.9053}$ \\
 & $\pm$ 0.0048 & $\pm$	0.0017 \\ %
\hline 
\end{tabular}
    \caption{Results on Weiz dataset}
    \end{subtable}%
         \begin{subtable}{0.50\columnwidth}
    \centering
    \noindent\setlength{\tabcolsep}{1pt}
\begin{tabular}{c|r|rr} 
\hline 
\centering
Method & ACC & NMI \\ [0.5ex]
\hline 
\centering

LRR & 0.4862 & 0.4297\\
RSC & 0.3485  & 0.3252 \\ 
OSC & 0.5931 & 0.4393\\
SSC & 0.3858 & 0.3137\\
LSR & 0.4548 & 0.4894\\
TSC & 0.4781 & 0.7129\\
\hline
TSS(M) & 0.5395 & 0.8049 \\
LTS(M) & 0.5509 & 0.8226 \\ 
MTS(M) & 0.6010 & 0.8270 \\
\hline
\bf Ours & ${\bf 0.7864}$ & ${\bf 0.8325}$ \\
& $\pm$0.0069 & 	$\pm$0.0041 \\
\hline 
\end{tabular}
    \caption{Results on Keck dataset}
    \end{subtable}%
\begin{subtable}{0.50\columnwidth}
    \centering
    \noindent\setlength{\tabcolsep}{1pt}
\begin{tabular}{c|r|rr} 
\hline 
Method & ACC & NMI \\ [0.5ex]
\hline 
LRR & 0.4051 & 0.4162\\
RSC &  0.3664   & 0.1881 \\ 
OSC & 0.6877 & 0.5846\\
SSC & 0.4998 & 0.4389\\
LSR & 0.4322 & 0.5183\\
TSC &  0.5340  & 0.7593\\
\hline
TSS(W) & 0.5944 & 0.7878 \\ 
LTS(M) & 0.6299 & 0.8128 \\ 
MTS(M) & 0.6433 &  0.8239 \\
\hline
\bf Ours & ${\bf 0.8700}$ & {\bf 0.8256}\\
& $\pm$ 0.0022 & 	$\pm$ 0.0018 \\
\hline 
\end{tabular}
    \caption{Results on UT dataset}
    \end{subtable}
\begin{subtable}{0.50\columnwidth}
    \centering
    \noindent\setlength{\tabcolsep}{1pt}
\begin{tabular}{c|r|rr} 
\hline 
Method & ACC & NMI \\ [0.5ex]
\hline 
LRR & 0.2249 & 0.2397\\
RSC & 0.3730   & 0.3418 \\ 
OSC & 0.5589 & 0.4327\\
SSC & 0.4758 & 0.3817\\
LSR & 0.3667 & 0.3979 \\
TSC & 0.5556 & 0.7721\\
\hline
TSS(K) & 0.5792 & 0.8286 \\
LTS(U) & 0.5980 & 0.8211 \\
MTS(U) & 0.6163 & 0.8314 \\ 
\hline
\bf Ours& ${\bf 0.8297}$ & ${\bf 0.8471}$ \\
& $\pm$ 0.0042 & 	$\pm$  0.0028 \\ 
\hline 
\end{tabular}
    \caption{Results on MAD dataset}
    \end{subtable}
    \label{tab:comparisons}
    \end{table*}

\section{Experimental results}
\label{sec:results}

\subsection{Datasets}
We used four human motion datasets to evaluate our proposed model, which we detail next. 
\\\noindent{\bf Weizmann action dataset (Weiz).}
The Weiz dataset \cite{gorelick2007actions} consists of 90 video sequences, each with a different person performing 10 different types of actions in an outdoor
environment. 
\\\noindent{\bf Keck gesture dataset (Keck).}
The Keck dataset \cite{lin2009recognizing} consists of 168 video sequences with 4 people performing 14 military gestures with perturbations in the background. The same gesture is repeated three times by each person. Each person is treated as a motion segmentation task.
\\\noindent{\bf UT interaction dataset (UT).}
The UT dataset \cite{ryoo2009spatio} consists of 20 videos, each containing 6 classes of human-human interactions. Each video sequence is around 60 seconds long.
\\\noindent{\bf Multi-model Action Detection dataset (MAD).}
The MAD dataset \cite{huang2014sequential} consists of 40 sequences captured from 20 subjects. Each subject performs 35 activities (e.g., jumping, kicking) continuously in two different indoor environments, each corresponding to a different sequence.
Each person is treated as a motion segmentation task.

 \begin{figure}[t]
\centering
\includegraphics[height=118mm,width=0.99\linewidth]{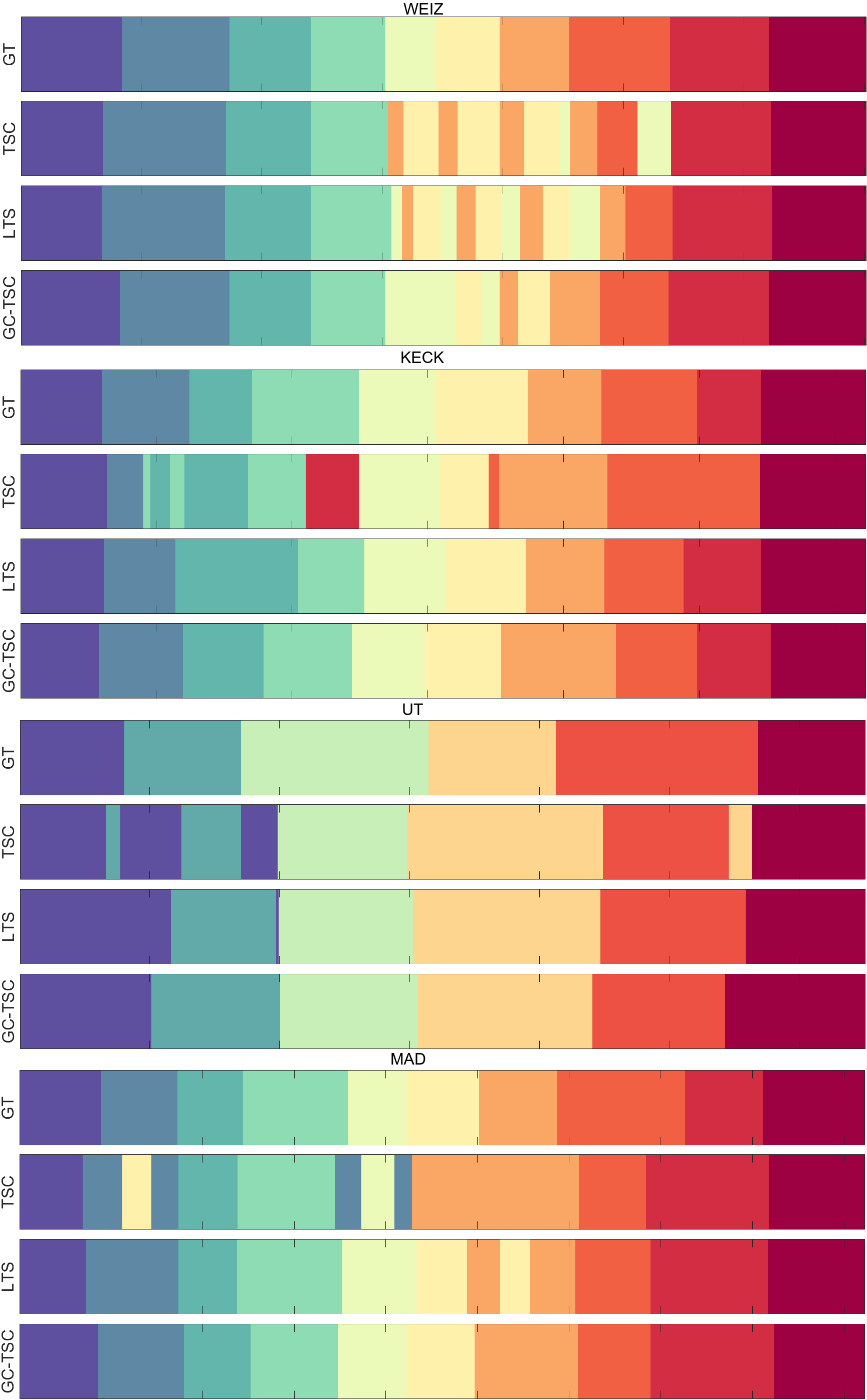}
\vspace{-2mm}
\caption{Clustering results on all datasets (sequences 1). Each color correponds to a cluster. GT is the ground truth.}
\label{fig:clusters}
\vspace{-2mm}
\end{figure}

\subsection{Experimental setup}

\noindent{\bf Dataset.}
We used the $n=324$ dimensional HOG features, extracted frame by frame and concatenated together, publicly available for these benchmark datasets\footnote{https://github.com/wanglichenxj/Low-Rank-Transfer-Human-Motion-Segmentation}, and used the same setting as in \cite{wang2018low}. 

\noindent{\bf Evaluation metrics and parameters settings.}
In all the experiments, we use the clustering accuracy
(ACC) and normalized mutual information (NMI) as the
evaluation metrics \cite{wu2009adapting}. The clustering accuracy is defined as 
$$\textnormal{ACC} = {\sum}_{i=1}^N \delta(g_i, map(c_i))/N,$$ 
where $\delta(\cdot,\cdot)$ is the Kronecker delta function and $map(c_i)$ is the \HWW{best} permutation mapping function which maps each cluster label 
\HWW{$c$ to a ground truth label $g$.}
The NMI is defined as 
$$\textnormal{NMI}(\Omega,\mathcal{C}) = \frac{I(\Omega, \mathcal{C})}{(H(\Omega) + H(\mathcal{C})/2}, $$
where $I$ denotes the mutual information and $H$ the entropy, $\Omega$ is the set of clusters and $\mathcal{C}$ is the set of classes. Higher values indicate better performance for the two metrics. 

We set the similarity bandwidth parameter $h$ in \eqref{equ:SC0} empirically to $h=0.0015$, which leads to visually satisfactory results for the graph affinity matrices ${G}$.
The hyperparameters $(\lambda_0,\lambda_1,\lambda_2)$ of our method are tuned independently for each dataset by grid search. Sensitivity to hyperparameter tuning is reported in Section \ref{sec:modelstudy} below.

\subsection{Comparative results}
We compare to several state-of-the-art subspace clustering based methods such as LRR \cite{liu2012robust}, SSC \cite{elhamifar2013sparse}, LSR \cite{lu2012robust}, temporal data clustering approaches such as TSC \cite{li2015temporal}, OSC \cite{tierney2014subspace},
and more recent 
Transfer Subspace Clustering based methods including Transfer
Subspace Segmentation (TSS) \cite{wang2018learning},  Low-rank Transfer Subspace (LTS) \cite{wang2018low}, Multi-mutual transfer subspace learning (MTSL) \cite{zhou2020multi}. For the latter class of methods, we show in Tab. \ref{tab:comparisons} the best results obtained independently of the source data as reported in \cite{wang2018low} and \cite{zhou2020multi}. 
We computed the best performance on the four benchmark datasets using the method \cite{li2019robust} that we named Robust Subspace Clustering (RSC), and report average values over five random seeds obtained by using the code made available by the authors \footnote{https://github.com/YuanmanLi/github-MWEE}. 
Bold indicates the best performance. 
Our method consistently obtains considerably better performance, also when compared to the transfer subspace clustering-based methods (TSS, LTS, MTS).
Qualitative results are shown in Fig.~\ref{fig:clusters}, where we show the estimated clusters for the first sequences of the four datasets, respectively. 
It can be appreciated that the segmentation obtained with the proposed approach is highly consistent and well aligned with the ground truth. Results obtained with TSC and LTS \footnote{https://github.com/wanglichenxj/Low-Rank-Transfer-Human-Motion-Segmentation} are less satisfactory and prone to produce spurious erroneous short segments, in particular close to segment transitions (e.g., at the center of the Weizmann dataset sequence), as well as merge segments into one cluster (e.g., at the center of the MAD dataset sequence). 

\begin{figure}[t]
    \centering
    \includegraphics[width=\linewidth,height=70mm]{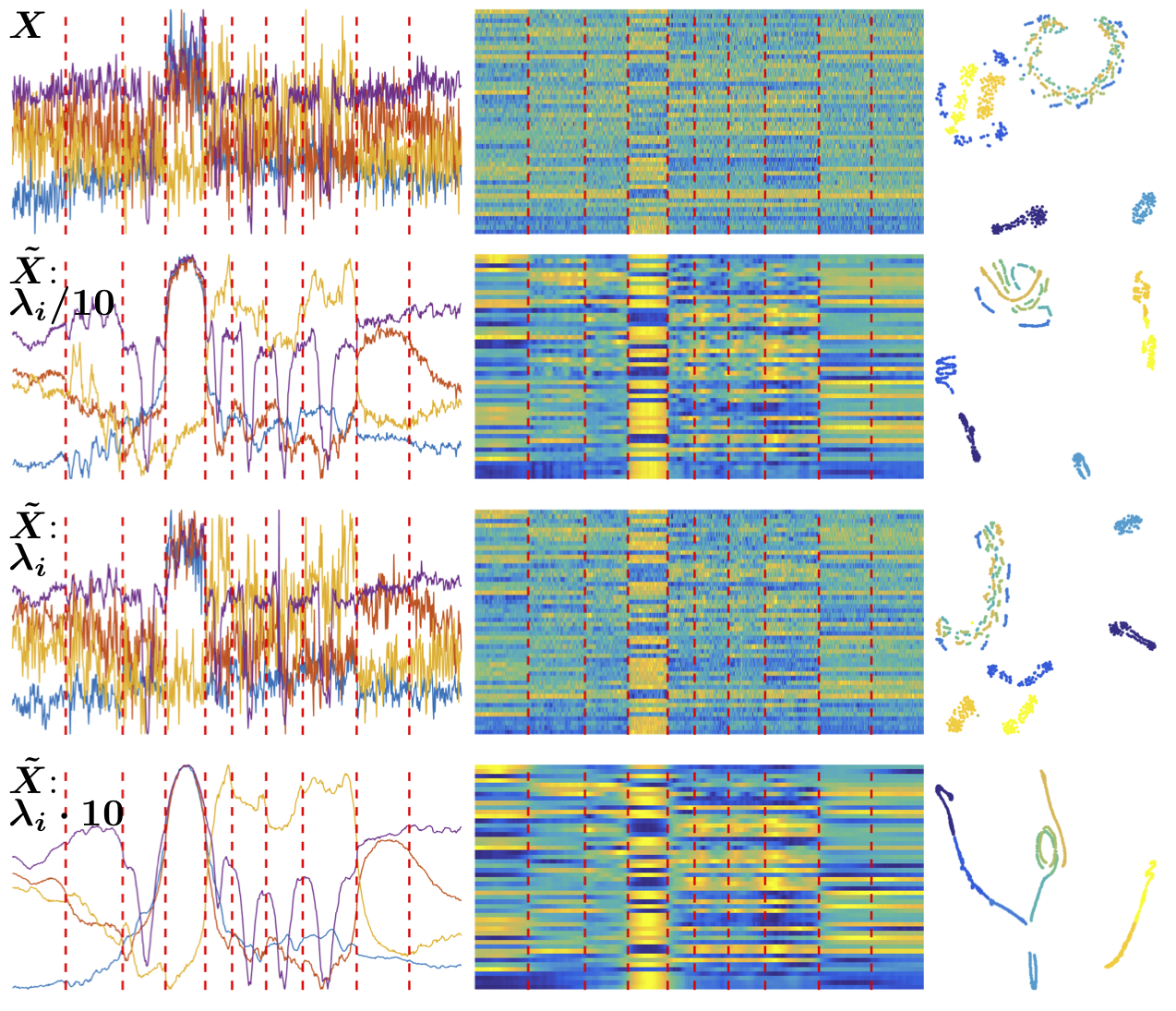}
    \vspace{-5mm}
    \caption{Illustration of $\tilde{X}$ for different weight for the graph penalty (from top to bottom). Left: arbitrary set of 4 features, Center: first 50 features. Right: t-SNE visualization.  }
    \label{fig:denoising}
\end{figure}

\begin{figure}[!ht]
\centering
\includegraphics[width=0.49\linewidth]{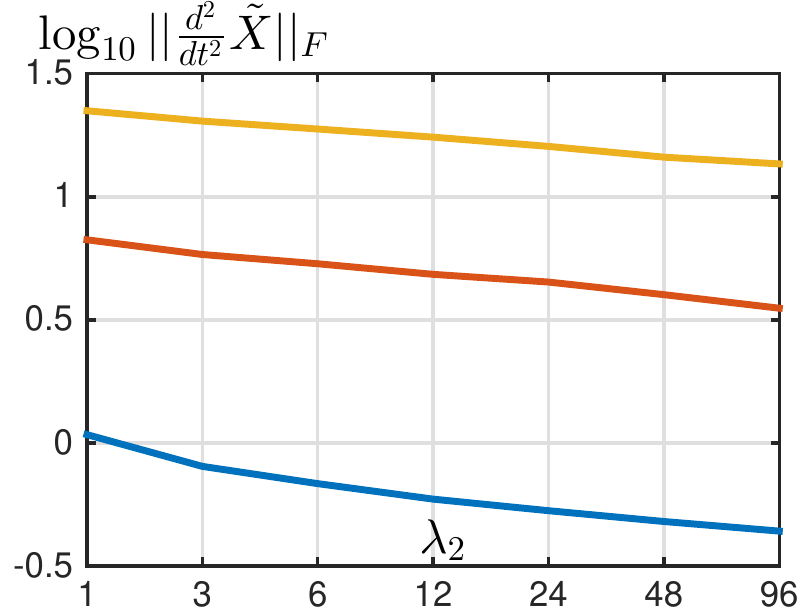}%
\includegraphics[width=0.49\linewidth]{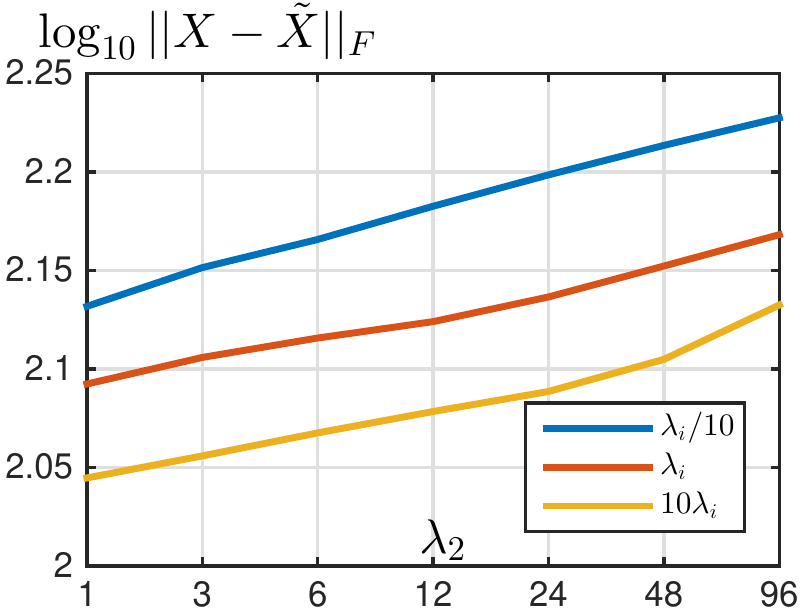}
\vspace{-2mm}
\caption{Frobenius norm of the temporal second derivative (left) and difference with data $X$ (right) of auxiliary representation $\tilde X$ for different values of the regularization term $\mathcal{L}_G$ and temporal regularization $\lambda_2$.}
\label{fig:Weiz1fi}
\vspace{-2mm}
\end{figure}

\begin{figure}[!ht]
    \centering
    \includegraphics[width=0.49\linewidth]{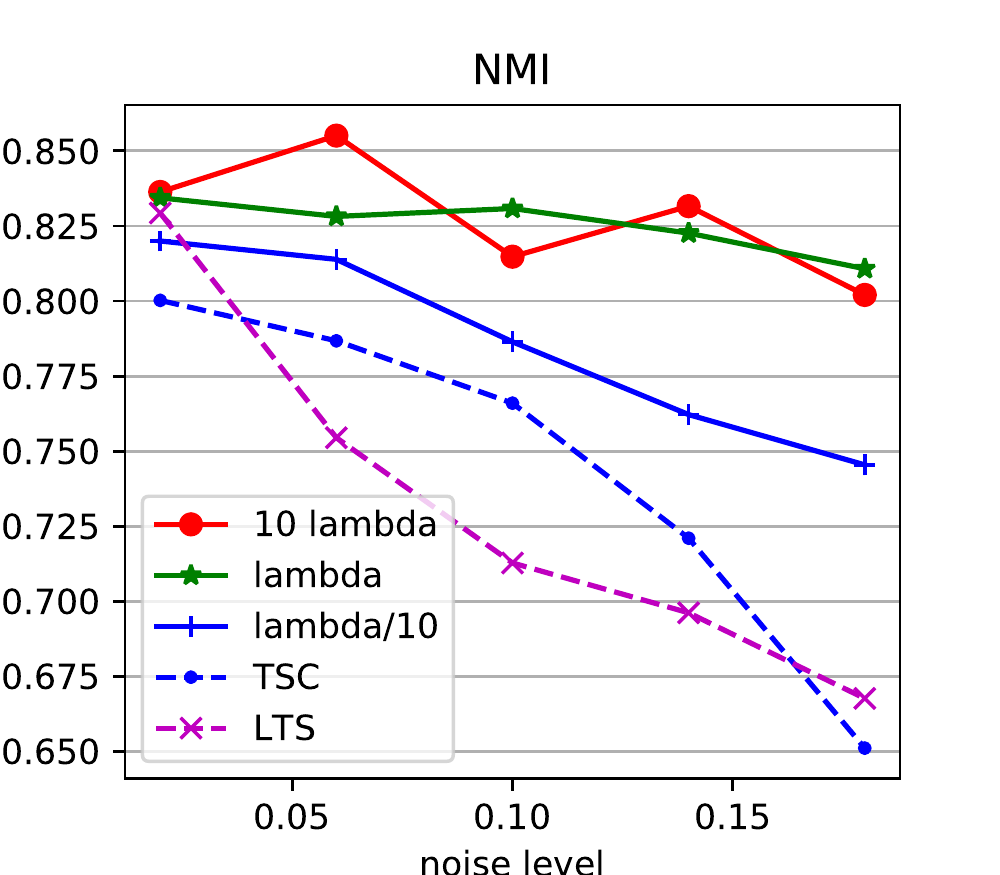}%
    \includegraphics[width=0.49\linewidth]{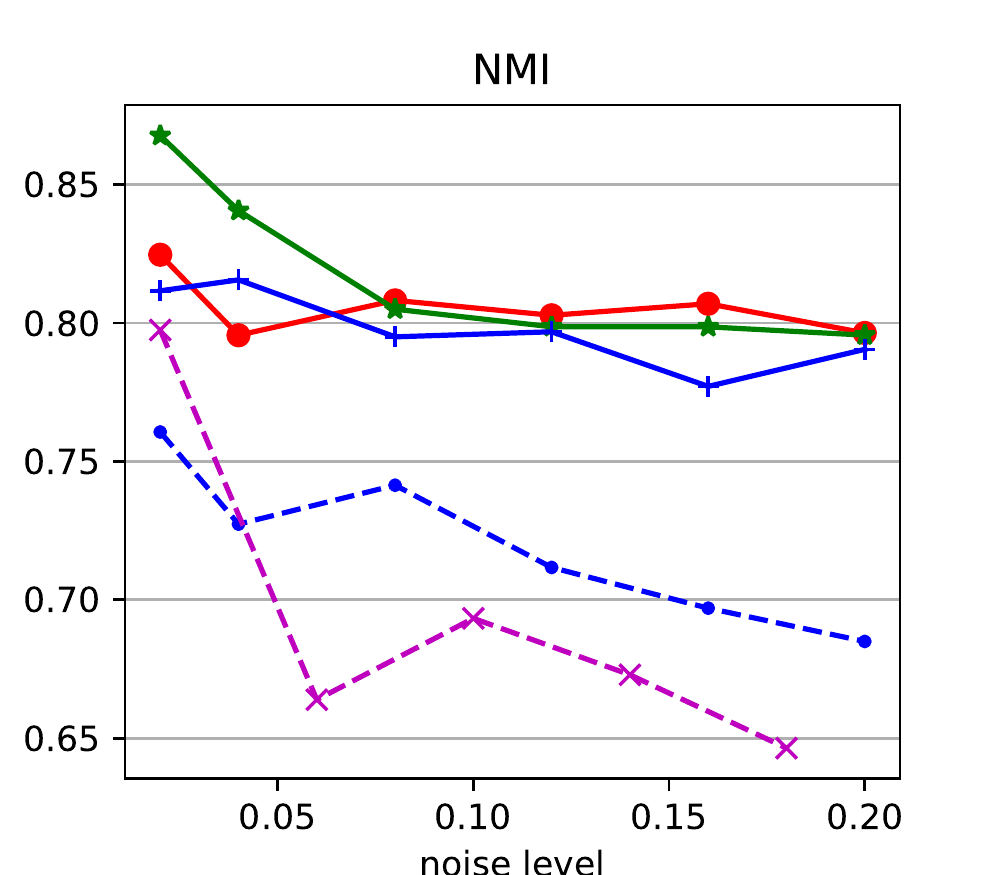}%
    \vspace{-2mm}
    \caption{Clustering performance (Weiz, subject 1) upon addition of centred piece-wise iid Gaussian noise to each frame, as a function of noise level: fixed piece-wise geometry (left) and randomly varying piece-wise geometry (right).}
    \label{fig:denoising:2}
    \vspace{-2mm}
\end{figure}

\begin{figure*}[t]
\centering
\includegraphics[height=0.185\linewidth, trim=15 0 52 0, clip]{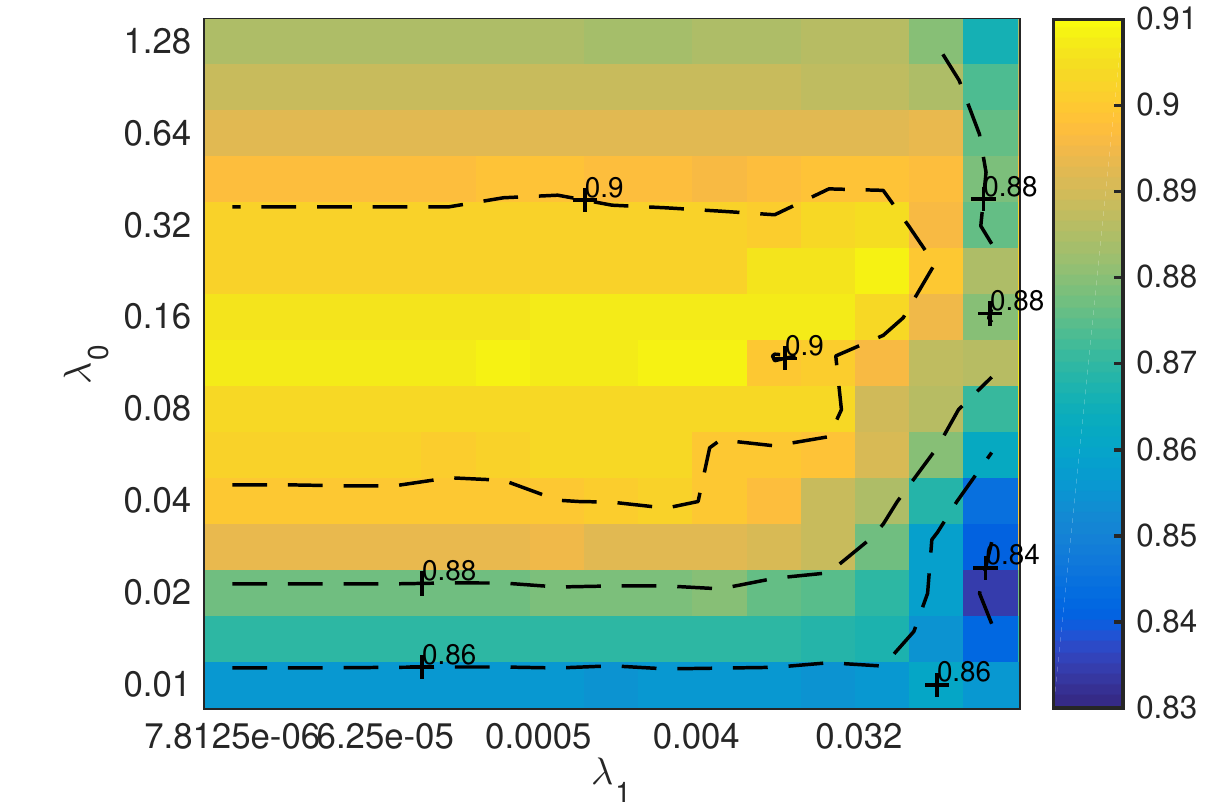}%
\includegraphics[height=0.185\linewidth, trim=0 0 52 0, clip]{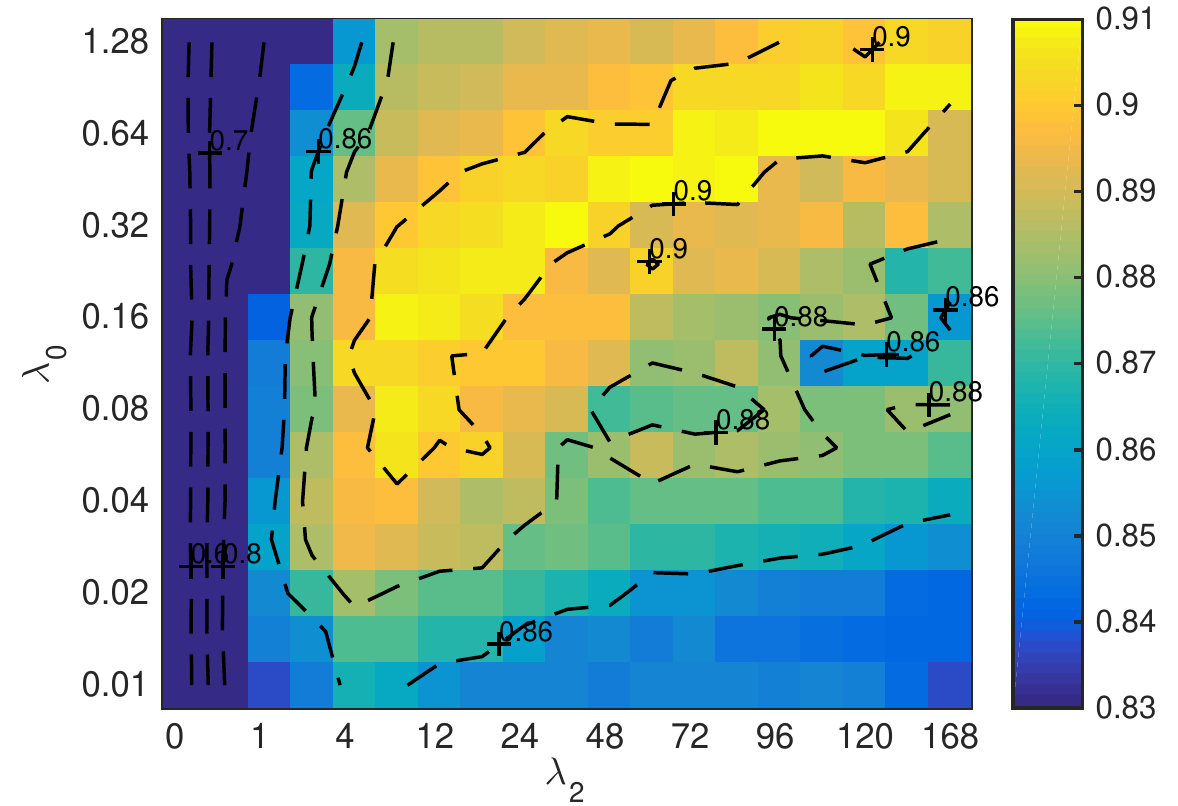}%
\includegraphics[height=0.185\linewidth]{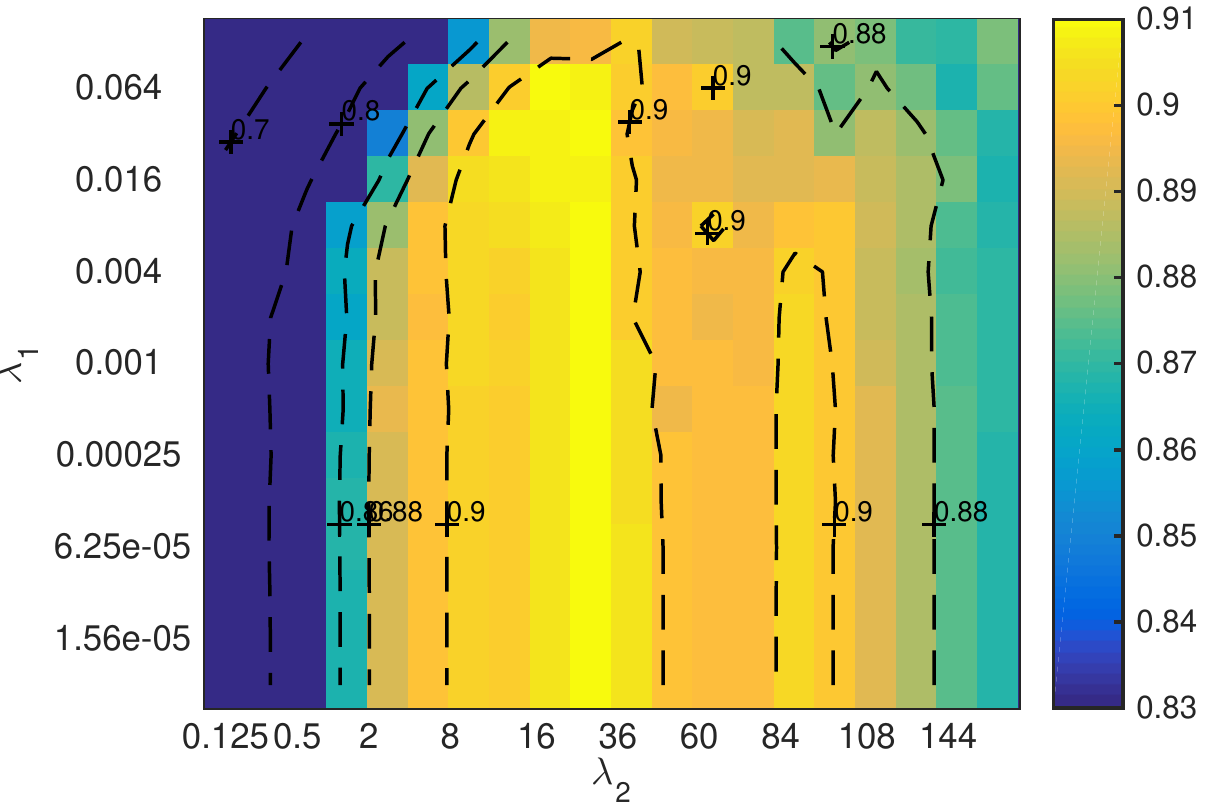}%
\includegraphics[height=0.185\linewidth]{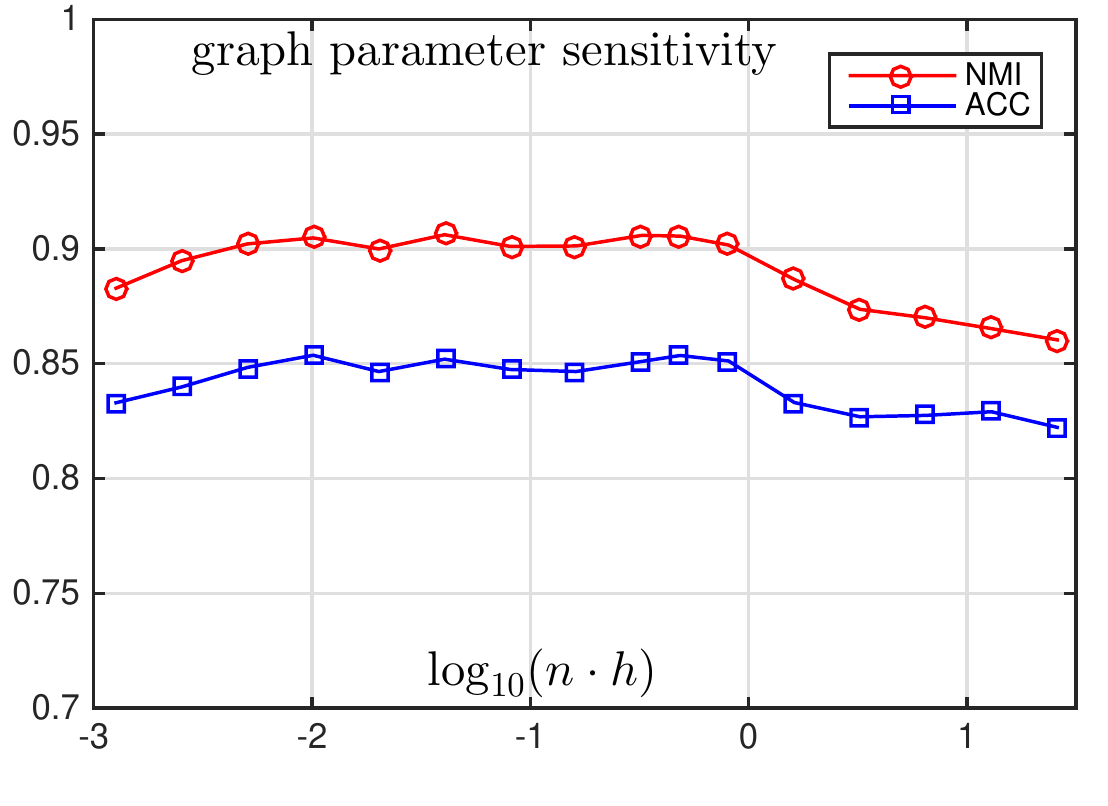}
\vspace{-3mm}
\caption{\label{fig:sensitivity}Sensitivity analysis for $\lambda_0$, $\lambda_1$, $\lambda_2$ in \eqref{equ:augLagNew0}, and for the graph filtering parameter $h$.}
\end{figure*}


\subsection{Model study}
\label{sec:modelstudy}

\noindent{\bf Graph constrained representation.}
We illustrate how our method leverages on the local geometric structure of the original data $X$ to produce an improved auxiliary data representation $\tilde X$.
Specifically, we compare, for the first subject of the Weizmann dataset, the data $X$ and the graph constrained data representation $\tilde{X}$ for different relative weight of the graph penalty $\mathcal{L}_G$ in \eqref{equ:augLagNew0} (realized by multiplying/dividing $\lambda_i,i=0,1,2$, by a common factor). In Fig.~\ref{fig:denoising} we plot a subset of 4 arbitrarily chosen features of $X$ and $\tilde{X}$ (plotted as time series, left plots), the 50 first features of $\tilde{X}$ (center plots) and  a visualization of $\tilde{X}$ by t-SNE \cite{van2008visualizing} (right plots); the vertical bars indicate true motion boundaries. 
It can be observed that the features $\tilde{X}$ resemble de-noised versions of the original data $X$, that become smoother and smoother as the relative weight of $\mathcal{L}_G$ in \eqref{equ:augLagNew0} is decreased and thus $\tilde X$ is more free to move from $X$, while at the same time retaining the global motion feature evolution and, in particular, sharp transitions at motion boundaries (cf., e.g., motion segment 4).  
The t-SNE visualization corroborates such observations and reveals that the clusters get more and more clearly separated in the high dimensional space.
 
\noindent{\bf Temporal versus Graph regularization.}
\HW{Fig.~\ref{fig:Weiz1fi} provides a more quantitative illustration of the effect of the graph penalty $\mathcal{L}_G$ (results obtained for the first subject of the Weizmann dataset).
Specifically, we plot the norm of the temporal second derivative of $\tilde{X}$ as a measure of smoothness (left plot) and of the norm of $X-\tilde X$ as a measure of data fidelity (right plot) as functions of temporal regularization parameter $\lambda_2$, and for different relative weight of $\mathcal{L}_G$ in \eqref{equ:augLagNew0} (blue, red and yellow color). 
The results for this experiment indicate that reducing the relative weight of $\mathcal{L}_G$ leads to auxiliary representations $\tilde X$ that are smoother and differ more from $X$, as expected.  \HWW{The same holds true for increasing $\lambda_2$. However, it can be appreciated that reducing the relative weight of the graph penalty term (i.e., letting $\tilde X$ move more freely)}
is more effective in controlling the temporal smoothness of $\tilde X$ \HWW{than tuning $\lambda_2$}, while at the same time retaining a significantly tighter fit with the original data $X$. For instance, decreasing the weight of $\mathcal{L}_G$ by a factor 10 increases smoothness by a factor $10^{0.7}\approx5-8$ (left plot) at only $10^{0.05}\approx10-15\%$ increased deviation from the original data, while increasing $\lambda_2$ by two orders of magnitude leads to modest smoothness increase of $10^{0.3}\approx2$ at more than $10^{0.08}\approx20\%$ larger deviation of $\tilde X$ from $X$.}

\noindent{\bf Noise modeling.} 
\HW{In Fig.~\ref{fig:denoising:2} we illustrate on the first subject of the Weizmann dataset the effectiveness of our model in dealing with noise. Specifically, we add centred Gaussian noise of varying global noise level (standard deviation) to the frames of the sequence before HOG feature extraction. Moreover, the noise variance \HWW{within each frame} is modeled piece-wise constant according to:
\HWW{i) a fixed geometry (staircase from left to right image boundary, results reported in Fig.~\ref{fig:denoising:2} (left)) and ii) quadrants whose size changes randomly from one frame to the other, emulating temporal burstiness of feature noise (Fig.~\ref{fig:denoising:2} (right)).}
Results (NMI) are given for three relative weights for the graph penalty $\mathcal{L}_G$ and compared with TSC and LTS and indicate that the graph constrained data representation $\tilde X$ effectively absorbs parts of the data corruptions and leads to significantly better clustering performance than the TSC and LTS approach that operate directly on the corrupted data $X$.}

\noindent{\bf Parameter sensitivity analysis.}
In our approach, three key regularization parameters, i.e., $\lambda_0$, $\lambda_1$ and $\lambda_2$, need to be manually tuned.
To study sensitivity to their variation on the model output, we fix the value of one parameter and vary the other two parameters over 2 to 3 orders of magnitude. Results for the Weizmann dataset are shown in Fig. \ref{fig:sensitivity} (left). Clearly, very small values for the temporal regularization weight $\lambda_2$ worsen the performance. This is to be expected because data temporal coherence is not modeled any longer when $\lambda_2\to0$. Otherwise, the results are consistent in a large range of parameter values. In particular, our proposed method obtains better NMI performance when $\lambda_0 \in [0.1,1.2]$ and when $\lambda_2\geq 5$.
Moreover, similarly to the observations in \cite{li2015temporal}, the performance of our model is found to vary little with the parameter $\lambda_1$ that regularizes the block-diagonal constraint.  Most importantly,  these results demonstrate that every term in our model is useful for improving performance.

Moreover, Fig.~\ref{fig:sensitivity} (right) 
shows the performance of our method for different values of the filtering parameter $h$ of the similarity kernel $S$ that tunes the data graph representations and hence the coupling of the original and the auxiliary data graphs $G_0$ and $\tilde G$. Segmentation performance are found to remain nearly constant over more than two orders of magnitude for $h$, which demonstrates that our approach is robust also to the precise choice of value for this parameter.

\begin{figure}
    \centering
    \includegraphics[width=0.5\linewidth]{UT_LagrFrob.pdf}%
    \includegraphics[width=0.5\linewidth]{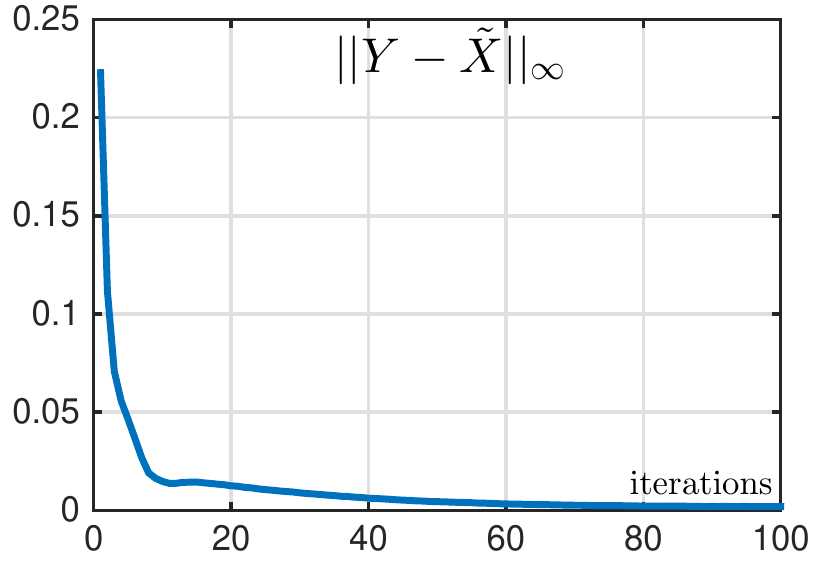}%
    \vspace{-3mm}
    \caption{Convergence analysis:  $||Y-\tilde X||_F$ (left) and $||Y-\tilde X||_\infty$ (right) as a function of iteration number (average over UT dataset for 5 seeds).}
    \label{fig:convergence}
\end{figure}

\begin{table}[]
    \centering\small
    \begin{tabular}{|l|r|r|r|r|r|}
     \hline
      NMI &   Weiz\hfill &  Keck\hfill &  UT\hfill &  MAD\hfill \\
        \hline
        \multirow{2}{*}{\bf Ours} 
         & 0.9053  & 0.8325 & 0.8256 & 0.8471 \\
         &  $\pm$ 0.0017 &  $\pm$ 0.0041  & $\pm$ 0.0018  & $\pm$ 0.0009\\
        \hline
        \multirow{2}{*} 
       {}Ours w/o  & 0.8277  & 0.8000 & 0.8176  & 0.7529  \\
      $\tilde{X}$, $\mathcal{L}_G$ &  $\pm$ 0.0028 & $\pm$ 0.0084  & $\pm$ 0.0056 & $\pm$ 0.0028\\
         \hline
    \end{tabular}
    \vspace{-3mm}
    \caption{Ablation study results. }
    \label{tab:ablation:large}
    \vspace{-2mm}
\end{table}

\noindent{\bf Convergence analysis.}
As the problem is non-convex, the global convergence of ADMM cannot be guaranteed theoretically. However, following \cite{xu2012alternating} the convergence property of ADMM can be shown if the first-order necessary conditions are satisfied. As we iterate over alternating the minimization of $U,V,D,Z$ with $\tilde{X}$ fixed, and the minimization of $Y,\tilde{X}$, the convergence of the algorithm is guaranteed if each of these minimization steps converges. The convergence of $U,V,D,Z$  with $\tilde{X}$ fixed has been proved in \cite{li2015temporal}. The convergence of the minimization of the graph regularization term is guaranteed by our design choice: the cross-entropy is a strongly convex function with respect the variables $Y$ and $\tilde{X}$,  $\tilde{X}$ is always positive since we normalize the features before performing gradient descent and therefore the mild conditions for the convergence of this term are satisfied. We show the convergence curve that demonstrates the convergence of our algorithm
(i.e., the update of the Lagrange multiplier associated to the graph regularization term over time)  in Fig. \ref{fig:convergence}.

\noindent{\bf Complexity analysis.}
 The most time-consuming parts in our model during optimization is the update of $V$ in TSC that has complexity $\mathcal{O}(r^2N)$. Denoting by $K$  the number of iterations at convergence, the overall computational complexity of our algorithm is  $\mathcal{O}(Kr^2N)$, thus equal to TSC.

\noindent{\bf Ablation study.}
To validate the effectiveness of the proposed model, we compare the performance of our method with that obtained when the graph based regularization term is removed and $\tilde{X}=X$ is fixed in the fidelity term. This corresponds to the TSC model of \cite{li2015temporal}, but with different values for the parameters $\lambda_i,i=0,1,2$, individually optimized here for each dataset. Quantitative results are reported in Tab. \ref{tab:ablation:large} for all four datasets%
\footnote{Results obtained using 5 random seeds and grid search to achieve best performance in terms of NMI, which explains quantitative differences with results from \cite{wang2018low,zhou2020multi} shown in Tab.\ref{tab:comparisons}.}
and confirm that our approach is highly effective and leads to consistent and significant improvements (NMI increase of up to +9\%).

\section{Conclusions}
\label{sec:conclusions}
We proposed a novel subspace clustering based approach for HMS which jointly learns an auxiliary data representation, a non-negative dictionary and a coding matrix under temporal, block-diagonal and graph constraints via an original ADMM formulation.
The rationale underlying our approach is to 
\HWW{operate on}
auxiliary data $\tilde{X}$ instead of the original time-series data $X$, lending the model extra flexibility for finding more expressive codes $Z$, while preserving  the  local geometrical structure of the original data in high-dimensional space.
We present a comprehensive analysis of our model, including several illustrations of the model behaviour,  model design justification, ablation study, parameter sensitivity and convergence analysis. 
Furthermore, we achieve  significant  performance  improvements of up  to $\approx 20\%$ (accuracy) and $\approx 5\%$ (NMI) over  the  state-of-the-art methods on four public benchmarks for HMS.

\section{Acknowledgements}
Work partially funded by projects MINECO/ERDF RyC, PID2019-110977GA-I00, JAEINT19-EX-0014, RED2018-102511-T. 
 
{\small
\bibliographystyle{ieee_fullname}
\bibliography{egbib}

\begin{thebibliography}{10}\itemsep=-1pt

\bibitem{babacan2012probabilistic}
S~Derin Babacan, Shinichi Nakajima, and Minh Do.
\newblock Probabilistic low-rank subspace clustering.
\newblock In {\em Advances in Neural Information Processing Systems}, pages
  2744--2752, 2012.

\bibitem{barbivc2004segmenting}
Jernej Barbi{\v{c}}, Alla Safonova, Jia-Yu Pan, Christos Faloutsos, Jessica~K
  Hodgins, and Nancy~S Pollard.
\newblock Segmenting motion capture data into distinct behaviors.
\newblock In {\em Proceedings of Graphics Interface 2004}, pages 185--194.
  Canadian Human-Computer Communications Society, 2004.

\bibitem{boyd2011distributed}
Stephen Boyd, Neal Parikh, and Eric Chu.
\newblock {\em Distributed optimization and statistical learning via the
  alternating direction method of multipliers}.
\newblock Now Publishers Inc, 2011.

\bibitem{chen1999motion}
William Chen and Shih-Fu Chang.
\newblock Motion trajectory matching of video objects.
\newblock In {\em Storage and Retrieval for Media Databases 2000}, volume 3972,
  pages 544--553. International Society for Optics and Photonics, 1999.

\bibitem{dimitrova1995motion}
Nevenka Dimitrova and Forouzan Golshani.
\newblock Motion recovery for video content classification.
\newblock {\em ACM Transactions on Information Systems (TOIS)}, 13(4):408--439,
  1995.

\bibitem{elhamifar2009sparse}
Ehsan Elhamifar and Ren{\'e} Vidal.
\newblock Sparse subspace clustering.
\newblock In {\em 2009 IEEE Conference on Computer Vision and Pattern
  Recognition}, pages 2790--2797. IEEE, 2009.

\bibitem{elhamifar2013sparse}
Ehsan Elhamifar and Rene Vidal.
\newblock Sparse subspace clustering: Algorithm, theory, and applications.
\newblock {\em IEEE transactions on pattern analysis and machine intelligence},
  35(11):2765--2781, 2013.

\bibitem{gholami2017probabilistic}
Behnam Gholami and Vladimir Pavlovic.
\newblock Probabilistic temporal subspace clustering.
\newblock In {\em Proceedings of the IEEE Conference on Computer Vision and
  Pattern Recognition}, pages 3066--3075, 2017.

\bibitem{gorelick2007actions}
Lena Gorelick, Moshe Blank, Eli Shechtman, Michal Irani, and Ronen Basri.
\newblock Actions as space-time shapes.
\newblock {\em IEEE transactions on pattern analysis and machine intelligence},
  29(12):2247--2253, 2007.

\bibitem{goyal2018graph}
Palash Goyal and Emilio Ferrara.
\newblock Graph embedding techniques, applications, and performance: A survey.
\newblock {\em Knowledge-Based Systems}, 151:78--94, 2018.

\bibitem{huang2014sequential}
Dong Huang, Shitong Yao, Yi Wang, and Fernando De~La~Torre.
\newblock Sequential max-margin event detectors.
\newblock In {\em European conference on computer vision}, pages 410--424.
  Springer, 2014.

\bibitem{jhuang2013towards}
Hueihan Jhuang, Juergen Gall, Silvia Zuffi, Cordelia Schmid, and Michael~J
  Black.
\newblock Towards understanding action recognition.
\newblock In {\em Proceedings of the IEEE international conference on computer
  vision}, pages 3192--3199, 2013.

\bibitem{keogh2001locally}
Eamonn Keogh, Kaushik Chakrabarti, Michael Pazzani, and Sharad Mehrotra.
\newblock Locally adaptive dimensionality reduction for indexing large time
  series databases.
\newblock In {\em Proceedings of the 2001 ACM SIGMOD international conference
  on Management of data}, pages 151--162, 2001.

\bibitem{keogh2003need}
Eamonn Keogh and Shruti Kasetty.
\newblock On the need for time series data mining benchmarks: a survey and
  empirical demonstration.
\newblock {\em Data Mining and knowledge discovery}, 7(4):349--371, 2003.

\bibitem{li2015temporal}
Sheng Li, Kang Li, and Yun Fu.
\newblock Temporal subspace clustering for human motion segmentation.
\newblock In {\em Proceedings of the IEEE International Conference on Computer
  Vision}, pages 4453--4461, 2015.

\bibitem{li2019robust}
Yuanman Li, Jiantao Zhou, Xianwei Zheng, Jinyu Tian, and Yuan~Yan Tang.
\newblock Robust subspace clustering with independent and piecewise identically
  distributed noise modeling.
\newblock In {\em Proceedings of the IEEE/CVF Conference on Computer Vision and
  Pattern Recognition}, pages 8720--8729, 2019.

\bibitem{lin2009recognizing}
Zhe Lin, Zhuolin Jiang, and Larry~S Davis.
\newblock Recognizing actions by shape-motion prototype trees.
\newblock In {\em 2009 IEEE 12th international conference on computer vision},
  pages 444--451. IEEE, 2009.

\bibitem{liu2012robust}
Guangcan Liu, Zhouchen Lin, Shuicheng Yan, Ju Sun, Yong Yu, and Yi Ma.
\newblock Robust recovery of subspace structures by low-rank representation.
\newblock {\em IEEE transactions on pattern analysis and machine intelligence},
  35(1):171--184, 2012.

\bibitem{lu2018subspace}
Canyi Lu, Jiashi Feng, Zhouchen Lin, Tao Mei, and Shuicheng Yan.
\newblock Subspace clustering by block diagonal representation.
\newblock {\em IEEE transactions on pattern analysis and machine intelligence},
  41(2):487--501, 2018.

\bibitem{lu2012robust}
Can-Yi Lu, Hai Min, Zhong-Qiu Zhao, Lin Zhu, De-Shuang Huang, and Shuicheng
  Yan.
\newblock Robust and efficient subspace segmentation via least squares
  regression.
\newblock In {\em European Conf. on Computer Vision}, pages 347--360. Springer,
  2012.

\bibitem{lv2006recognition}
Fengjun Lv and Ramakant Nevatia.
\newblock Recognition and segmentation of 3-d human action using hmm and
  multi-class adaboost.
\newblock In {\em European Conference on Computer Vision}, pages 359--372.
  Springer, 2006.

\bibitem{oh2008learning}
Sang~Min Oh, James~M Rehg, Tucker Balch, and Frank Dellaert.
\newblock Learning and inferring motion patterns using parametric segmental
  switching linear dynamic systems.
\newblock {\em International Journal of Computer Vision}, 77(1-3):103--124,
  2008.

\bibitem{pan2009survey}
Sinno~Jialin Pan and Qiang Yang.
\newblock A survey on transfer learning.
\newblock {\em IEEE Transactions on knowledge and data engineering},
  22(10):1345--1359, 2009.

\bibitem{ryoo2009spatio}
Michael~S Ryoo and Jake~K Aggarwal.
\newblock Spatio-temporal relationship match: Video structure comparison for
  recognition of complex human activities.
\newblock In {\em 2009 IEEE 12th international conference on computer vision},
  pages 1593--1600. IEEE, 2009.

\bibitem{seeger2004gaussian}
Matthias Seeger.
\newblock Gaussian processes for machine learning.
\newblock {\em International journal of neural systems}, 14(02):69--106, 2004.

\bibitem{shi1998motion}
Jianbo Shi and Jitendra Malik.
\newblock Motion segmentation and tracking using normalized cuts.
\newblock In {\em Sixth International Conference on Computer Vision (IEEE Cat.
  No. 98CH36271)}, pages 1154--1160. IEEE, 1998.

\bibitem{smyth1999probabilistic}
Padhraic Smyth.
\newblock Probabilistic model-based clustering of multivariate and sequential
  data.
\newblock In {\em Proceedings of the Seventh International Workshop on AI and
  Statistics}, pages 299--304. San Francisco, CA: Morgan Kaufman, 1999.

\bibitem{tierney2014subspace}
Stephen Tierney, Junbin Gao, and Yi Guo.
\newblock Subspace clustering for sequential data.
\newblock In {\em Proceedings of the IEEE conference on computer vision and
  pattern recognition}, pages 1019--1026, 2014.

\bibitem{van2008visualizing}
Laurens Van~der Maaten and Geoffrey Hinton.
\newblock Visualizing data using t-sne.
\newblock {\em Journal of machine learning research}, 9(11), 2008.

\bibitem{vidal2011subspace}
Ren{\'e} Vidal.
\newblock Subspace clustering.
\newblock {\em IEEE Signal Processing Magazine}, 28(2):52--68, 2011.

\bibitem{wang2018learning}
Lichen Wang and Zhengming Ding.
\newblock Learning transferable subspace for human motion segmentation.
\newblock In {\em AAAI}, 2018.

\bibitem{wang2018low}
Lichen Wang, Zhengming Ding, and Yun Fu.
\newblock Low-rank transfer human motion segmentation.
\newblock {\em IEEE Transactions on Image Processing}, 28(2):1023--1034, 2018.

\bibitem{wu2015ordered}
Fei Wu, Yongli Hu, Junbin Gao, Yanfeng Sun, and Baocai Yin.
\newblock Ordered subspace clustering with block-diagonal priors.
\newblock {\em IEEE transactions on cybernetics}, 46(12):3209--3219, 2015.

\bibitem{wu2009adapting}
Junjie Wu, Hui Xiong, and Jian Chen.
\newblock Adapting the right measures for k-means clustering.
\newblock In {\em Proceedings of the 15th ACM SIGKDD international conference
  on Knowledge discovery and data mining}, pages 877--886, 2009.

\bibitem{xia2017human}
Guiyu Xia, Huaijiang Sun, Lei Feng, Guoqing Zhang, and Yazhou Liu.
\newblock Human motion segmentation via robust kernel sparse subspace
  clustering.
\newblock {\em IEEE Transactions on Image Processing}, 27(1):135--150, 2017.

\bibitem{xiong2002mixtures}
Yimin Xiong and Dit-Yan Yeung.
\newblock Mixtures of arma models for model-based time series clustering.
\newblock In {\em 2002 IEEE International Conference on Data Mining, 2002.
  Proceedings.}, pages 717--720. IEEE, 2002.

\bibitem{xu2012alternating}
Yangyang Xu, Wotao Yin, Zaiwen Wen, and Yin Zhang.
\newblock An alternating direction algorithm for matrix completion with
  nonnegative factors.
\newblock {\em Frontiers of Mathematics in China}, 7(2):365--384, 2012.

\bibitem{yang2010temporal}
Yun Yang and Ke Chen.
\newblock Temporal data clustering via weighted clustering ensemble with
  different representations.
\newblock {\em IEEE transactions on knowledge and data engineering},
  23(2):307--320, 2010.

\bibitem{zhou2012hierarchical}
Feng Zhou, Fernando De~la Torre, and Jessica~K Hodgins.
\newblock Hierarchical aligned cluster analysis for temporal clustering of
  human motion.
\newblock {\em IEEE Transactions on Pattern Analysis and Machine Intelligence},
  35(3):582--596, 2012.

\bibitem{zhou2020multi}
Tao Zhou, Huazhu Fu, Chen Gong, Jianbing Shen, Ling Shao, and Fatih Porikli.
\newblock Multi-mutual consistency induced transfer subspace learning for human
  motion segmentation.
\newblock In {\em Proceedings of the IEEE/CVF Conference on Computer Vision and
  Pattern Recognition}, pages 10277--10286, 2020.

\end{thebibliography}
}

\end{document}